\documentclass{article}

\usepackage{arxiv}

\usepackage[utf8]{inputenc} 
\usepackage[T1]{fontenc}    
\usepackage{hyperref}       
\usepackage{url}            
\usepackage{booktabs}       
\usepackage{amsfonts}       
\usepackage{nicefrac}       
\usepackage{microtype}      
\usepackage{lipsum}
\usepackage{graphicx}
\graphicspath{ {./images/} }

\title{A new solution and concrete implementation steps for Artificial General Intelligence}

\author{
 Yongcong Chen \\
New Millennium Future Technology Co., LTD.\\
 Beijing, China\\
  \texttt{yongcongchen@sina.com} \\
   \And
 Ting Zeng\\
 Library, Tsinghua University\\
  Beijing, China\\
  \texttt{ceng-t@mail.tsinghua.edu.cn} \\
  \And
 xingyue Chen\\
  High School Department of Qinghe Middle School\\
  Haidian District, Beijing, China\\
 \texttt{} \\
}

\begin{document}
\maketitle
\begin{abstract}
In this paper, we propose a new approach to building a artificial general intelligence with self awareness, which includes: (1) a new method to implement attention mechanisms; (2) a way to give machines self-demands; (3) how to form a value evaluation system compatible with the network; (4) a way to create the world models; (5) how to realize a top-down, hierarchical thinking decision-making chain; (6) a way to achieve general decision-making and response capabilities; (7) a way for a machine to directly obtain human experience through language. In the paper, we first analyze some of the shortcomings of current LLMs (Large Language Model) and propose ideas for improvement. Then we analyze why our scheme can solve the above problems and provide detailed steps for implementing our scheme. In chapter 4, we have presented a step-by-step mplementation roadmap. And in  chapter 5, we have presented a specific implementation demonstration. In chapter 6, we analyze the advantages and disadvantages of our scheme and propose further research directions. 
\end{abstract}


\section{Introduction}
At present, multi-modal large models are moving towards universality\cite{ref1}\cite{ref35}\cite{ref36}\cite{ref36}\cite{ref37}. However, there is still a final step to reach truly universal artificial intelligence (AGI), which is difficult for large models to cross. This step is that currently, AI has difficulty correctly decomposing complex tasks into a series of simple tasks and handling unexpected situations. The root of the problem lies in the fact that current AI can only receive "reward" feedback after tasks are completed. When facing a large number of optional intermediate processes, if it cannot see the final reward, it cannot make the correct choice\cite{ref7}. To solve this problem, the two existing technical routes are: (1) Try all the main intermediate processes to establish decision-making knowledge. This is "reinforcement learning." (2) Or add some rules to string tasks together. This is "Agent." These rules can either come from human programming or be given by language large models, but in essence, they all come from statistics. \cite{ref8}\cite{ref9}\cite{ref10}\cite{ref11}\cite{ref12}\cite{ref3}

The core defects of the above schemes are: (1) Many things cannot be done with "reinforcement learning," such as having robots care for the elderly. (2) Rules can never be added completely. When changing tasks or environments, the machine needs to "relearn." Therefore, we propose a new implementation path: adding "self-needs" to the machine. Let it, like humans, make decisions according to "seeking advantages and avoiding disadvantages and satisfying self-needs" regardless of the task. Therefore, compared with current technologies, what we are creating is a "person," while what large models create is a "tool." In fact, we believe that there is no "universal" tool in the world. The only thing that can be universal at present is "people". Therefore, to create AGI, a feasible path is to create human-like intelligence.

\section{Characteristics of human-like intelligence}
To create human-like intelligence, we need to analyze the similarities and differences between human intelligence and current AI intelligence.
 First, we note that the information that humans can recognize is only a very small part of the information in our world. This is because humans have limited resolution for information. Therefore, in the process of evolution, based on their own resolution capabilities, humans have produced Tokens. Therefore, Tokens are the smallest information units commonly used by humans. 
 
For humans, from the Big Bang to today, we have been living in a four-dimensional information matrix composed of Tokens (a tensor of three-dimensional space + time dimension). This four-dimensional Tokens matrix contains all the information that humans can obtain. Through memory and forgetting mechanisms, humans gradually summarize those common Token combinations and use some language symbols (voice, pictures, or symbols) to represent them. This is a concept. Humans use concepts to describe any information in the matrix. This is chatting and writing articles using language. 

Obviously, the language symbol itself is only a component of the concept, and the concept represents a common Token combination. A concept is a set of coordinate basis clusters in the information space matrix we are in. Under such a basis cluster, the coefficient expression of common information is sparse. This is because humans, for the convenience of expressing common information, have formed a set of basis clusters agreed upon within the group through memory and forgetting mechanisms and survival of the fittest by summarizing common Token combinations. Then, using such a basis cluster, any spatial and temporal arrangement of Tokens is expressed. Obviously, Chinese and English are two sets of basis clusters in the information space. They can both be used to express any information in the matrix. \cite{ref38}\cite{ref25}\cite{ref26}\cite{ref27}\cite{ref28}

In deep learning, the coefficients of each layer of neural network are behind a group of basis clusters that do not appear explicitly. From layer A neural network to layer A + 1 neural network, in essence, it is a process of basis transformation from the A basis cluster to the A + 1 layer basis cluster. Then, through a nonlinear function, part of the information is compressed or discarded. Then the next attempt is made. Therefore, the essence of deep learning is to regard all input information as a matrix. Then, using the trial method, find a suitable matrix basis. When using this basis to decompose or generate the information of concern, the coefficient matrix is sparse. 

Obviously, humans do not have the powerful computing power of computers. They rely on memory and forgetting mechanisms to summarize common Token combinations and use symbols to represent them. Humans take common combinations as the basis and have redundancy and can only efficiently express common information. While deep learning simultaneously calculates all the information of concern and finds the optimal basis cluster. Therefore, when the basis cluster found by deep learning expresses the overall information of concern, it is more efficient. But the basis cluster found by deep learning is too different from the human basis cluster, making it difficult for the two to communicate. 

The essence of the attention mechanism is to increase the weight of common Token combinations and prioritize using them as the basis to decompose and generate information (vectors)\cite{ref39}\cite{ref17}\cite{ref18}\cite{ref19}. Assuming a complete basis cluster of the information space is X, and its dimension is N. If we project all input vectors (A, B, C...) to the subspace K of X and also project all input vectors to the subspace Q of X. Then, compare each component of vector A in subspace K with all components of all input vectors (A, B, C...) in subspace Q one by one (implemented by dot product). If there are similar components between the component of vector A in subspace K and the component of vector A in subspace Q, it indicates that these components have a higher probability of occurrence and may belong to common components. Therefore, the weight of these components should be increased. If there are similar components between the component of vector A in subspace K and the component of vector B in subspace Q, it can be considered that there is a connection relationship between A and B through this component (the vector distance is closer). Therefore, the weight of these components should be increased in the input vectors. 

The components of each vector can be regarded as an attribute of the vector. A 512-dimensional word vector can be regarded as 512 coefficients obtained by projection on the implied 512 dimensions. Therefore, when the weight of a component is increased, it means that the weight of the attribute corresponding to this component has increased in the current application environment. This is similar to when humans use concepts. In different application environments, the weights of each attribute of the same concept will change. 

Obviously, through the above attention mechanism, we obtain the pairwise correlation weights of each input information vector and all other vectors. Similarly, we can take the vector with modified component weights as the new input, project it to the new subspaces K' and Q' again, and obtain their combined correlation again by comparing the components one by one. Obviously, this method can be iterated to form more levels of combined correlation. 

Obviously, in each projection process to the subspace K or Q of X, we can also change it to project to the sub-spaces K1, K2, K3 and Q1, Q2, K3 of X. There is no essential difference. And this process is exactly the implementation process of the multi-head attention mechanism in Transformer. 

Furthermore, similar components widely existing in different input vectors indicate that they are common components and represent the commonalities of a certain type of thing. That is, they are knowledge obtained through information compression (dimensionality reduction). Obviously, this is a statistical process. Due to the endless combinations of language, this is a non-complete statistical process. In Transformer, the statistical result is the weight matrix. The reasoning process of Transformer is to use the known local statistical correlation to infer the correlation between components under a specific input vector combination. This is a Bayesian process. Transformer, through the attention mechanism, obtains the basis cluster of the information space, which is similar to the basis cluster created by humans through memory and forgetting mechanisms and survival of the fittest. Therefore, Transformer that combines deep learning and attention mechanism realizes finding the basis like human habits. This is the real reason for the emergence of intelligence in large models. Language large models are like aliens decrypting human language: At the beginning, when the sample is not enough, the "basis cluster" it extracts is very different from the human "basis cluster," so it will constantly make mistakes, grope in the dark, and always hit walls. As the number of samples continues to increase, there is a higher probability that its "basis cluster" will align with the human "basis cluster." But this is not a linear process. For example, before reaching a certain threshold, groping in the dark makes little progress. At a certain node, if the accuracy rate reaches the threshold, the entire decryption process will be greatly accelerated and completed quickly. This is the "emergence" phenomenon. What emerges in the machine is not intelligence but finding the correct basis cluster. 
However, facing the requirements of "AGI," large models have insurmountable difficulties: 

1.Large models do not retain the original spatiotemporal order of Token combinations.

First, although Transformer has position encoding, large models use deep learning. The "basis cluster" corresponding to each layer is implicit, while the corresponding coefficient matrix is explicit. Obviously, large models only create or generate "common combinations" without retaining all the input information itself. This will make it difficult for large models to overcome the hallucination problem. 
Second, although both use common tokens combinations as the basis, there is a translation process between the basis of large models and the basis of humans. The basis cluster of large models is the implied basis cluster corresponding to the coefficients in its latent space. The translation process from these basis clusters to the content generation of large models is established through a large amount of training data. For language, the cost of establishing this translation process is still acceptable. But for other modalities, especially those involving interactive decision-making and action sequences, the cost is high or even impossible to achieve. For example, when taking care of infants and caring for the elderly, we cannot obtain corner cases data because we cannot use people to do various extreme experiments. 

It is because large models do not retain the original spatiotemporal order of token combinations that humans find it difficult to understand the knowledge formed by large models. As a result, humans cannot imitate its knowledge structure, preset innate knowledge for the machine, or modify the knowledge created by the machine. This brings two serious problems: (a) Innate knowledge cannot be preset. This makes it impossible for us to create a machine baby and then let her learn like a human baby. (b) Knowledge cannot be updated in real time.

2.Large models do not synchronously generate value chains and need to be remedied by reinforcement learning or Agent after the fact, making it difficult to handle corner cases. 

During the pre-training process of large models, no value chain is formed. Instead, it is remedied after the fact through Reinforcement Learning from Human Feedback (RLHF). This kind of remedy is difficult to cover corner cases and lacks safety. So, for the above two problems, our solutions are: 
1.Make the bases of the two completely consistent and remove the translation process of the basis cluster.

2. Synchronously generate the value chain during the training stage.

\section{A Path to Realize Human-like Intelligence }
\subsection{Brief description}
First, give a brief introduction to the steps to implement our scheme, and then give a specific description for each step.
Step 1. Tokenize information. (Same as other AI technologies)

Step 2. Tensorize tokens. (Establish a memory bank)

Step 3. Propagate activation values from input tokens to tokens in the memory bank according to similarity relationships. 

Step 4. For all activated tokens, propagate activation values to adjacent tokens according to proximity relationships. 

Step 5. For each activated token, propagate activation values in a chain in the memory bank according to the principles of similar activation and proximity activation.
Among them, in Step 3 to Step 5, the higher the similarity, the greater the transfer coefficient. The closer the storage location, the greater the transfer coefficient. The higher the memory value of a token, the greater the transfer coefficient. 

Step 6. Accumulate the activation values obtained by each token from different propagation paths. 

Step 7. The activation values of all tokens fade over time. 

Among them, Step 3 to Step 7 are the chain association activation process. This is the reasoning process of the attention mechanism. The activation value is the reasoning weight.

Step 8. Each token updates its memory value in a positive correlation according to the magnitude of its obtained activation value. Moreover, all memory values fade over time. Therefore, after each inference, the input information will be stored in the memory bank. And after each inference, the memory bank will be updated. Therefore, in our scheme, the machine will store "facts". The pre-training and reasoning of the machine are the same process. The machine is lifelong learning, and knowledge is updated in real time. This is closer to the human learning process.

The memory value of each token is its pre-training weight value. In memory, there are a large number of token combination methods. Those token combinations that can repeat appear. The tokens they contain can activate each other every time and push up each other's activation values, thereby obtaining higher memory values. In the memory bank, the local network composed of those tokens with high memory values is a "concept". And under specific input excitation, the activation value of the activated tokens in the "concept" is the weight of each component of the "concept" in the input scenario. This is completely equivalent to the reasoning process in Transformer. 

Through the above eight steps, we can find common "token combinations". Common "token combinations" include both spatial and temporal combination methods. They are "concepts", that is, "world models".

Therefore, our scheme and Transformer reach the same goal by different routes. They both implement the attention mechanism and find common "token combinations". The difference is that the common "token combinations" we find do not destroy their temporal and spatial order, so we can understand and imitate them. 

Step 9. Preset innate knowledge. Innate knowledge includes the lowest innate needs, rewards and punishments, emotions, and necessary innate safety instinct knowledge. Of course, other knowledge can also be preset. This knowledge is implemented by presetting tokens, as well as the time and space positions of tokens, and the memory values of tokens. They exist as part of the memory bank and seamlessly integrate with the acquired memory formed later to form an overall memory bank. 

"Fine-tuning" of innate knowledge can be achieved through supervised data or through trial and then manual modification. In addition, the acquired knowledge accumulated through later learning and innate knowledge jointly participate in decision-making. Therefore, innate knowledge does not need to be very accurate. 

Step 10. Let innate needs, rewards and punishments, and emotions (represented by special tokens), and acquired information (ordinary token information flow), form a time information flow in the training and life of the machine and be stored in chronological order. Then, through the chain association activation process + attention mechanism, a fully connected knowledge network (memory bank) is formed. 

With a fully connected knowledge network, when the machine faces any input, its decision is: pay attention to the currently activated reward tokens and punishment tokens. Then reduce the probability of occurrence of tokens that transmit activation values to punishment tokens, and increase the probability of occurrence of tokens that transmit activation values to reward tokens. Obviously, this is the first step in decision-making. 

Then, for those tokens that need to reduce the probability of occurrence or need to increase the probability of occurrence, it is necessary to reduce or increase the probability of occurrence of tokens that transmit activation values to them. Obviously, this is a top-down, layer-by-layer decomposition of the goal process. The implementation of this process can be achieved by using the optimal path search method and realizing it by counting the activation values of activated reward tokens or punishment tokens on different paths; it can also use a neural network to take the activated tokens, their memory values and their activation values as the encoding of the latent space and map them to the output sequence through deep learning. 

Different from large models, the encoding of our latent space can be understood and imitated. Therefore, the machine can directly obtain all the input-output correspondence relationships that have been formed in human history through language learning. That is to say, in our scheme, the machine can, like humans, obtain corresponding skills through classroom learning and a small number of experimental operations. Therefore, our scheme is human-like intelligence. 

At present, large models need to go through reinforcement learning. That is to say, they need to obtain corresponding skills through personal training in all scenarios. In fact, a more serious problem is that the current large model scheme requires the machine to simultaneously strengthen learning of all skills in a single training process in order to achieve universality. This is obviously impossible. 

\subsection{Implementation details}
Next, we will give further explanations for each step. In our scheme, each token is a data record. They are composed of the four fields shown in Table 1.
Table 1. Composition of each token data.
\begin{table}
 \caption{Composition of each token data.}
  \centering
  \begin{tabular}{llll}
    \toprule
    \cmidrule(r){1-2}
    Filed1     & Filed2     & Filed3  &Filed4 \\
    \midrule
    Record Time &Token & Memory value  & Activation value   \\
    \bottomrule
  \end{tabular}
  \label{tab:table}
\end{table}
Field1 is the time stamp, represents the time record when the token is stored in the memory bank. Field2 can be data from graphics, speech or other sensors. It represents the token itself and is a tensor. Field3 represents  the pre-trained weight and is a scalar. Field4 represents the inference weight of the attention mechanism and is a scalar. 

\subsection{Step 1. Tokenize information}
This step is the same as the existing multimodal AI scheme. The machine only needs to extract the tokens from the input information, such as the tokens corresponding to text, the tokens corresponding to voice, etc. For images, according to human habits, priority should be given to extracting low-resolution tokens corresponding to the whole. Then, the machine will process the obtained tokens and make the next response according to the decision. The next response includes whether to further extract tokens in the image, which areas of interest to extract, and what resolution to use.

Even if the token extraction algorithm is not perfect at the beginning, it doesn't matter. Because our scheme is to continuously summarize common tokens and common token combinations through memory and forgetting mechanisms. So the token extraction of the machine is a gradually optimized process. Through the memory and forgetting mechanism, survival of the fittest is achieved. Those widely existing tokens or token combinations will be retained and form more complex tokens. And those tokens that can rarely repeat will be eliminated and they will no longer be extracted. Therefore, the machine can find those widely existing tokens through training data and reverse-optimize the extraction algorithm according to the need to extract them. This is similar to humans. It is a gift brought to humans by evolution. Because extracting such underlying programs as tokens requires extensive reuse to achieve the maximum energy efficiency ratio, which is of great significance for evolution.

\subsection{Step 2. Matrixize tokens}
Each token corresponds to a record in the memory bank. It has four fields as shown in Table 1. The memory value indicates the memory intensity. If it is zero, it will be deleted. The activation value indicates the activation intensity. If it is zero, it means it is not activated. All records are stored by the simultaneity storage method, and spontaneously form the entire memory bank.

Regarding the simultaneity storage method, the specific implementation methods include:

(2.1) The machine retains the relative time positions of tokens appearing in the input information.
One implementation method is: the machine uses the distance of tokens in the storage space to reflect the time distance between the moments when these tokens are stored. For example, the machine stores tokens in turn according to the input time sequence. The closer the time, the closer the storage location; 
Another storage method that retains the relative time position is that each token has coordinates in the memory space; the coordinates in the memory space mainly include the storage time information of tokens;

(2.2) The machine retains the relative spatial positions of tokens appearing in the input information; one implementation method is: the machine overlays and places the tokens extracted each time with the original data according to the position, angle and size with the highest similarity to the original data, and retains the relative positions of these tokens in space when storing;

The implementation method can also be: extract overall low-resolution tokens first, and then extract other local tokens on demand according to the machine's decision. In this way, through the adjacent storage relationship, local tokens and overall tokens not only have an adjacent activation relationship, but also have a similarity relationship between tokens, so they will activate each other and establish a positional relationship connection.

\subsection{Step 3. Perform similarity activation from input tokens to tokens in the memory bank}

Give each input token a unified initial activation value A0. A0 itself is a preset value. But it can be adjusted by the activation value of the reward symbol and punishment symbol activated in the last machine chain association activation process.

The activation value of the activated reward symbol and punishment symbol is the machine's prediction of the potential reward and punishment value for the previous input information. And the initial activation value A0 will affect the range of the chain association activation process. When the initial activation value A0 is very high, the propagation range of the chain association activation process is larger. This is because in our scheme, the activation value propagation coefficient is less than 1. As the number of chain propagation stages increases, the propagated activation value becomes smaller and smaller. When the activation value obtained by a token is less than the preset threshold, the chain propagation process will terminate. So A0 reflects the machine's degree of attention to the input information. When A0 is very high, the machine will activate more tokens in memory to find memories related to input tokens. This is similar to humans. If the previously input tokens bring a high potential reward and punishment, then the input of new related tokens may be paid special attention. For example, the words of the boss will make you think of more information.

The principle of similarity activation is: (1) The higher the similarity between tokens, the greater the transfer coefficient; (2) The higher the memory value, the greater the transfer coefficient; the memory value is the pre-training weight. It needs to be emphasized that the same token may continuously appear in many positions in the memory bank. They all have their own different memory values. This is because under different token arrangements, the weight of the same token in it is not the same. This is similar to the attention mechanism in large models.

\subsection{Step4, Perform proximity activation for all activated tokens}

We believe that the proximity relationship between tokens represents that there is some implicit correlation between them. The closer they appear in time, the closer the potential relationship. This correlation can be statistically calculated through the chain association activation process + memory and forgetting mechanism. The proximity relationship actually reflects a token combination relationship. If this combination relationship can repeat appear, then it is a common combination. So we use the proximity activation process in the chain association activation process to utilize common combinations.

Each activated token will transmit activation values to its adjacent tokens; the closer the time position, the greater the transfer coefficient; the higher the memory value, the greater the transfer coefficient. In the memory bank, if there is a proximity relationship between tokens, it indicates that they were once a combination method. If their memory value is high, it indicates that they are a common combination method. If only one token has a high memory value, it means that they are not a common combination method. If the memory values of tokens are not high, then the activation values they propagate are very low, and the chain propagation of tokens stops quickly. This indicates that such information is not important and their weight in information processing is very low.

	Tokens use the method of "the closer the time position, the greater the transfer coefficient; the higher the memory value, the greater the transfer coefficient" to activate common combinations containing them. In essence, it is to calculate the correlation between the input token and common token combinations.

	If N input tokens all transmit activation values to X token combinations in memory that contain them, then these X combinations will obtain very high activation values. Because each token will simultaneously activate multiple tokens in the X combination according to similarity and proximity. Therefore, the X combination obtains a higher activation value through the way of accumulating activation values. These higher activation value tokens form a "model", which is the expected model activated by the input vector (N tokens).

	In essence, the chain association activation process is the process of decomposing the input vector into the coordinate base cluster. And the machine decides whether to further recognize information and what resolution to adopt and which part of the input information to recognize according to the activated reward and punishment information after decomposition. This is the process of pattern recognition of information.

\subsection{Step5, Chain Activation Process}
Each input token performs "similarity activation" and "proximity activation" in the memory bank, and the magnitude of activation value transfer is positively correlated with their pre-training weight (memory value).

	For each activated token in the memory bank, if their activation value exceeds the preset threshold, they also perform "similarity activation" and "proximity activation" in the same way. The magnitude of activation value transfer is positively correlated with their pre-training weight (memory value).

	This process is carried out in a chain until all input tokens complete their own "chain activation process". Therefore, in addition to activating the vectors in memory that are similar to the input vector, the machine will also activate the "causes" and "consequences" of the vectors in memory that are similar to the input vector, that is, the previous information and the subsequent information in time in the memory bank. And different memory fragments may activate different "causes" and "consequences".

\subsection{Step6, Accumulate activation values}
If a token in a certain memory bank has an activation value propagation path with multiple input tokens (that is, either directly related or indirectly related), the activation values transmitted from the input are accumulated. Therefore, tokens in the memory bank that have direct/indirect correlations with multiple input tokens will obtain higher accumulated activation values from multiple propagation paths.

	In this way, among the input tokens, if there are tokens that are related to each other, they will push up the weight of related tokens in the memory bank. That is to say, those common combinations, their activation values will rise from the activation value sea level. And this activation value sea level is the low activation value of those large numbers of tokens. Those tokens that rise from the activation value sea level constitute one or more "world models".

And those memories that are most relevant to the input, although they may not be common, due to their direct correlation with the input and short propagation paths, they may also obtain high activation values.

	Therefore, our scheme can obtain common models and pay attention to specific factual details. Therefore, our scheme comes with a "fact database" and can solve the "hallucination" problem of GPT at present.

\subsection{Step7, Activation values fade over time}
All activation values continuously decrease over time. When the tokens are input later, the relevant tokens in memory are activated. And the relevant tokens activated by the previous input have not completely faded, and the activation values will be accumulated.

	And the machine's decision is based on all activated tokens. Therefore, both the previous and subsequent input information will be taken into account.    
    Therefore, the thinking of the machine has a certain time coherence and can solve problems such as "omission", "reference", and "metaphor".
    
	Therefore, our machine utilizes the implicit relationship between the previous and subsequent inputs. This is the attention mechanism.

\subsection{Step8, Obtain Pre Training Weight Matrix}
In our scheme, those token combinations that can repeat appear may obtain higher memory values due to their repeatability. And because they are repeatable combinations, each time they push up each other's activation values, they obtain a memory value that is much higher than that of a simple single token repeatability. So this is a positive cycle process. So, from this we can see that our machine can summarize experience by itself. But at the same time, it will also be a time-consuming process to forget the existing thinking mode. Therefore, in our scheme, the pre-training statistical process of the machine is not a simple statistical process, but is completed jointly by the attention mechanism + memory and forgetting mechanism + the principle of seeking advantages and avoiding disadvantages.

	The decision-making principle of a machine is to seek advantages and avoid disadvantages. In the decision-making process of seeking advantages and avoiding disadvantages, the machine's recognition process of information is to selectively recognize information according to the way of seeking advantages and avoiding disadvantages. The machine will adjust the initial activation value A0 assigned to the input tokens according to the magnitude of "advantages and disadvantages" predicted by the previous decision. And the initial activation value A0 will affect the range and cumulative size of activation value propagation. This is to adjust the attention intensity according to "advantages and disadvantages". This is very similar to humans. At this point, it surpasses the current technology (Transformer). In fact, this is very similar to the human decision-making process. For example, the words of the boss can make you have more associations, activate more reward or punishment symbols, and thus more deeply predict the reward and punishment value.
Memory and forgetting mechanism: For all tokens in the memory bank, if they are activated once, their memory values will be updated in a positive correlation according to the magnitude of their activation values. Their memory values are the pre-trained weight matrix. Since the arrangement of tokens cannot be exhausted, this is a non-complete statistical process, similar to the pre-training process of large models.

The process of chained associative activation is essentially the process of projecting the input vector onto the coordinate basis cluster established by the attention mechanism. In our scheme, the activation value obtained through the chained associative activation process is the inference weight of the attention mechanism in Transformer. The latent space composed of the token combinations established by the chained associative activation process and their activation values is equivalent to the latent space formed after the input information is encoded by multiple layers of attention mechanisms in Transformer. The difference lies in that our encoding retains the original time and space order of tokens.

In fact, it is to know the probability of some token combinations and find the probability of a specific token combination. This is Bayesian inference implemented in the form of a neural network. In large models, the weight matrix represents the probability of known token combinations. In our scheme, the probability of known token combinations is represented by the positions of tokens and their memory values, and the inference probability is represented by the positions of tokens and their activation values.

It can be seen that the way we implement the attention mechanism is small-sample and cumulative learning. Moreover, the weight matrix is updated in real time, so the knowledge of our scheme is updated in real time. And we do not distinguish between the pre-training and inference processes, so our machine is lifelong learning.

It can be seen that our scheme does not need the BP algorithm, and its computational complexity is basically close to the inference process of large models. At the same time, in our scheme, supervised data can also be used and the BP algorithm can be used to fine-tune the memory values of tokens.
In addition, we can see that the chained associative activation process is highly patterned and can be directly implemented at the hardware level using new storage devices. This will help localize the calculations in our scheme, which will help expand the landing scenarios and reduce costs.

\subsection{Step 9, Preset Minimum Machine Requirements}
We have not only implemented the attention mechanism and found common token combinations but also have not disrupted the original time and space organizational form of tokens. Therefore, the knowledge network formed by our scheme is understandable to humans. So, we can imitate the organizational form of tokens in the final memory bank and establish the initial minimum innate memory for the machine. This is equivalent to presetting a minimum innate knowledge similar to human innate knowledge (the knowledge that babies are born with) for the machine.

In innate memory, it needs to contain the machine's minimum "demand system," "reward and punishment system," and "emotional system." The method is to use special tokens to represent each "demand," "reward and punishment," and "emotional state." Then, implant the minimum innate knowledge by imitating the final form of the memory bank (that is, the token arrangement + appropriate memory value).

Then, during the training process, let these special tokens representing "demand," "reward and punishment," and "emotional state" establish connection relationships with tokens representing the outside world (including the machine's own state parameters). Then, through the chained associative activation process + decision-making based on seeking advantages and avoiding disadvantages + memory and forgetting mechanism, the fittest survive and the common combinations of tokens are finally obtained. These common combinations are "common sense."

Common sense is the "world model," which contains the "world model" of human cognition of the external world and also contains the relationship between the "world model" established by humans and "me." It should be particularly noted that tokens are not only static features but also contain those simple dynamic features (such as rotation, swaying, etc.), so the world model is not static or fixed. The tokens that make up the "world model" will have different weights (activation values) under different input excitations.

The world model is related to the machine's training data and even the sequence of training data. This is similar to humans. A person's "world model" is related to his/her life experience.

With the world model, input tokens can activate those reward and punishment tokens, emotional tokens, and demand tokens through the chained associative activation process. And the activation value transfer path from input tokens to these characteristic tokens is a logical reasoning process compatible with neural networks. It is explicit, understandable, and can be imitated, so the machine's decision-making can be seen.
In fact, in the actual process of creating "general artificial intelligence," Step 9 is essentially the first step. But we can train experimental data through the previous steps to obtain and understand the organizational form of knowledge created by the machine. Then, by imitating these organizational forms, we can implement Step 9.

(1)Preset the basic demand advantage and disadvantage system related to the machine's life activities.
For example, give a reasonable range for power data. In "innate memory," preset a symbol representing "hunger." Put a "punishment" symbol and an emotional symbol representing "hunger" next to the "hunger" symbol. And give them appropriate memory values. When the power is insufficient, the life state monitoring program will directly give the initial activation value to the "hunger" symbol in "innate memory." Its activation value will spread chainwise throughout the entire memory bank. The emotional symbol of "hunger" next to it is activated, and the "punishment" symbol next to it will also be activated. So the machine has the emotion of "hunger" and a "punishment value" appears. To avoid the "punishment value," the machine will use its own experience and take the initiative to look for a plug to charge.

(2)Preset the advantage and disadvantage of "higher-order needs" of the machine's values. The simplest communication means need to be preset, and then values are cultivated.

Values need to be cultivated from an early age. So we need to cultivate the "values" of robots through education from an early age. Since education is needed, it needs to be achieved through "rewards" and "punishments." So at the very beginning, the machine needs to be able to recognize "rewards" and "punishments." Only in this way can we initiate the first step of learning through "rewards" and "punishments." Therefore, we need to imitate the organizational form of acquired memory networks and let the machine have innate knowledge that can recognize the simplest "rewards" and "punishments." For example, preset the most basic nodding feature (assuming X tokens) / shaking head feature (assuming Y features), without the need for precision.

Put a "being respected" symbol next to the nodding tokens. Put a "reward" symbol next to the "being respected" symbol. Give these symbols higher memory values so that the relationship between them becomes long-term memory. When some nodding tokens appear in the information input, through the chained associative activation process, the machine obtains a "reward value." In order to pursue the "reward value," the machine may plan various decisions in the future, with the purpose of obtaining "human nodding." Similar to a child, starting from the simplest communication method and gradually obtaining complex learning abilities, the logical chain of his/her gradually established "reward function" is: "milk" → "pacifier" → "feeding bottle" → "milk powder can"... →.... "academic performance" → "house and car"... → "social status".... → "life ideal." Therefore, after training, there are a large number of reward and punishment-related token symbols and token combinations closely related to these reward and punishment tokens in the machine's memory bank. There is a causal relationship between them. The things, behaviors, and results represented by these token combinations closely related to reward and punishment tokens are values. Therefore, any value of the machine can be established by presetting innate communication means and then cultivating step by step. In fact, humans are the same.

\begin{figure} 
    \centering
    \caption{The method To preset innate knowledge}
    \includegraphics[width=1\linewidth]{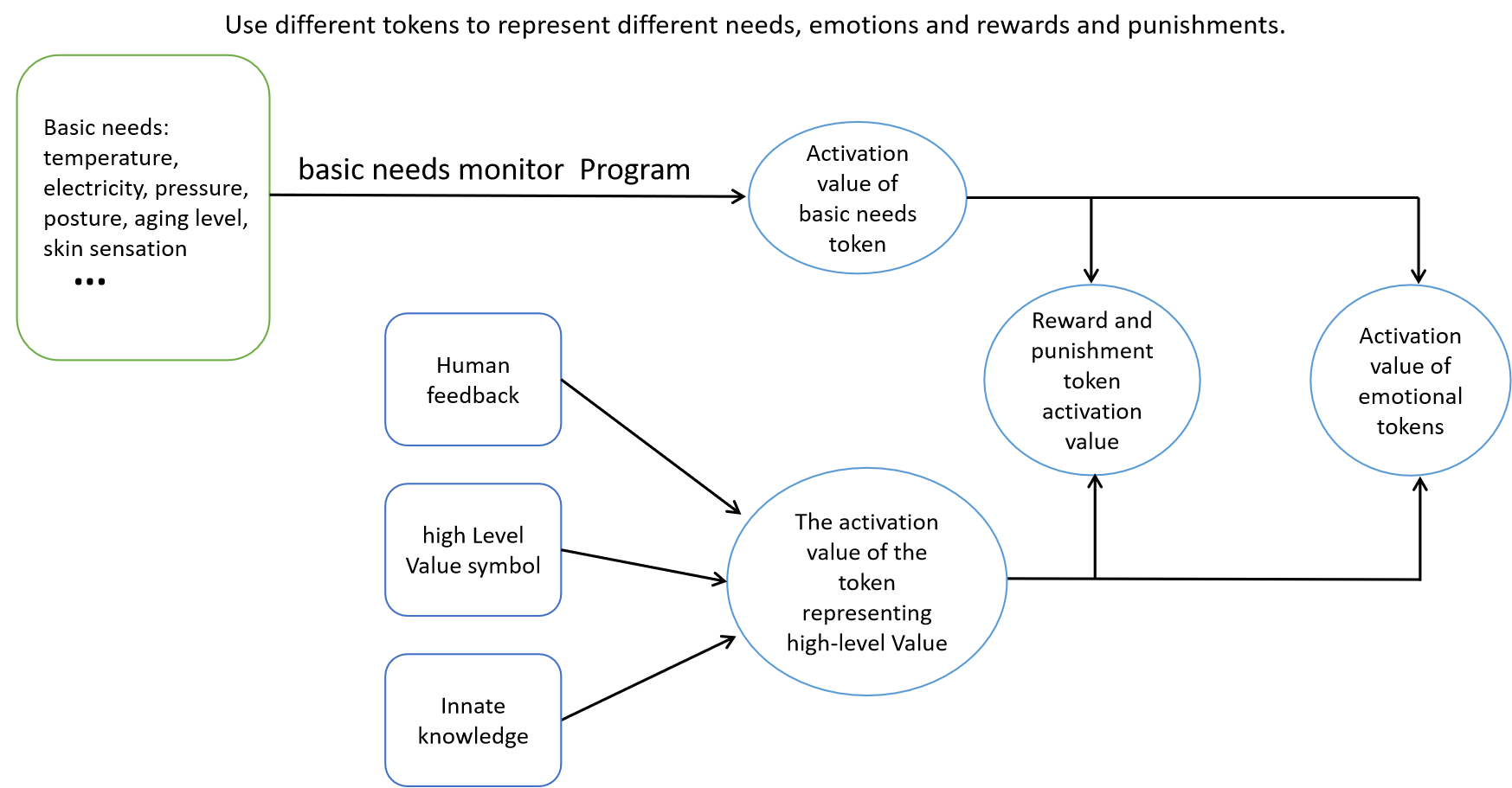}
\end{figure}

\subsection {Step 10, Form a Fully Connected Knowledge Network}
In our scheme, a network is finally formed in which each token is composed of four fields: time stamp, the token itself, memory value, and activation value.
	A large number of tokens are stored according to time intervals. By optimizing memory values and activation values, a knowledge network is formed. The memory value among them represents the pre-trained weight; the activation value among them represents the inference weight under the attention mechanism.
	Our network has both objective tokens and subjective tokens. The connection relationship formed by them through the attention mechanism is knowledge, and common knowledge is "common sense."
    
	This is the reason why our machine can predict advantages and disadvantages and make autonomous decisions. Because it has "needs" and a "value chain" related to "needs." The "value chain" is the activation value transfer path composed of tokens. Driven by needs, it will take the initiative to learn and self-iterate. For example, it will charge itself or go to the library to read books by itself.
	In our scheme, knowledge unfolds around "needs," and decision-making also unfolds around "needs." This is the core reason why our machine can achieve "generality." It faces only one task: "needs," rather than various "external tasks." Therefore, our scheme is an "active agent," while all other current schemes are "passive agents."
    
	It can be seen that our scheme is small-sample learning, knowledge is updated in real time, and the training and usage processes are integrated. Therefore, the machine is lifelong learning and self-iterating.
    
	Since the machine's knowledge exists in the form of a memory bank, and the memory bank is established through small-sample cumulative learning. Different memory banks can be directly spliced together to form a large memory bank, and the knowledge of two robots can be fused. Therefore, after the memory bank of a chef and the memory bank of a doctor are fused, the robot can have the skills of both a chef and a doctor without putting a large amount of data of chefs and doctors together for retraining. At present, other AI technical routes cannot achieve this. In large models, a large amount of data of doctors and chefs must be used for training at the same time before the machine can possibly master these two skills simultaneously. Obviously, according to such a training method, it is an extravagant hope to expect the machine to have "various" abilities.

\begin{table}
 \caption{Composition of each Token data.}
  \centering
  \begin{tabular}{llll}
    \toprule
    \cmidrule(r){1-4}
    Time mark    & Tokens     & Memory value  &Active value \\
    3001298 & the & 0.5000000  & 0.0392157   \\
    3001299	& machine & 0.3529412 & 0.9803922 \\    
    3001300	& can	& 0.7364326	& 0.9849463 \\    
    3001301	& follow	& 0.4730328	& 0.9460656 \\    
    3001302	& the	& 0.5000000	& 0.0392157 \\
    3001303	& preset	& 0.3745098	& 0.8843137\\    
    3001304	& instructions	& 0.6742372	& 0.3529412 \\    
    3001305	& and	& 0.5000000	& 0.0000000 \\    
    3001306	& use	& 0.8750000	& 0.9235294 \\    
    3001308	& input	& 0.6078431	& 0.3725490 \\    
    3001309	& information	& 0.5294118	& 0.1372549 \\
    \bottomrule
  \end{tabular}
  \label{tab:table}
\end{table}

Because the knowledge of the machine exists in the form of a memory database, and the memory database is stored in chronological order. What we want to keep is relative time interval information, not absolute time information. So different memory database can be directly stitched together to form large memory database. So, by combining the chef’s memory database and the doctor’s memory database, the robot can have the skills of both a chef and a doctor instantly, without having to retrain a lot of a chef and a doctor’s data together. The current AI technology solution, can not achieve this point. In LLM, it has to be trained with a large amount of both doctor and chef data for a machine to master both skills. Obviously, with such training method, it is a wild hope that a machines can have “all kinds of” abilities such as doctors, teachers, nannies and president.
Table 2  is a simple schematic table of memory database structure (Text tokens only ).

\section{Decision-making process}
\subsection{We have established a world model}
After obtaining new tokens each time, the machine needs to find paths to achieve rewards and avoid punishments based on the updated activation values. The set of these paths is the overall response path. The overall response path may be a network-like structure, and many local paths may lead to either reward symbols or punishment symbols.

	Due to the activation value transfer path leading to reward symbols (or punishment symbols), that is, we have achieved the pre-positioning of the reward and punishment function. Therefore, we have solved the problems of sparse and delayed reward functions in the current reinforcement learning process. Through a process similar to AlphaGo's optimal response path search, the machine can find the initial optimal response path.
	If the cumulative reward and punishment value of the whole does not enter the acceptable preset value (or does not converge), the machine cannot decide whether to select or exclude certain specific paths to maximize benefits. Then the machine needs to further recognize the input information and add more tokens to subdivide certain specific reward and punishment activation value transfer paths, thereby further helping the machine select or exclude certain specific paths. This step is a process spontaneously created by the machine to actively seek information to help decision-making. This process is iterated until the reward and punishment value statistics reach the acceptable preset value or converge.
    
	When further recognizing the input information, high-activation value tokens are either because their memory values are high, such as tokens representing a certain type of thing, or tokens closely related to the current input tokens, such as similarity or often appearing in close proximity. Therefore, the combination of high-activation value tokens activated in memory is a representative token combination related to the current input information. These representative token combinations are the "expected model" temporarily created by the machine. It comes from both the summary of past experience (the memory value of tokens after survival of the fittest) and is directly related to the current specific input. It is temporarily created through high activation values and is the "expected model" of the machine for the current combination of input tokens.
    
	The machine refers to the spatial or temporal relationship between tokens that have appeared in the input and tokens that have not appeared in the input in the "expected model", and takes the time and space positions of the tokens that have already appeared as the benchmark to predict the possible time or space positions where those tokens that have not yet appeared may appear. These high-activation value tokens that have not yet appeared in the expected model are expected tokens. The machine allocates the time and space positions of the sensor search of the machine according to the time, space, and size of the expected tokens in the expected model, and determines the sensor type to be used according to the attributes of the expected tokens (such as voice, image, or touch), and determines the required resolution according to the attributes of the expected tokens (such as size). This is the "on-demand recognition" process of the machine. This process can be iterated. This is a pattern recognition method of the machine for the world and can greatly save the computational effort of the machine.
    
	Selective attention is a means for extracting tokens from input information. The machine extracts tokens from the input information according to the recognition interval and resolution given by selective attention recognition. Only in this way can the problem of wireless granularity of image information be solved (the machine extracts information from images on demand). When the machine extracts data from a specific interval, in accordance with the principle of overall feature priority within the interval, it first extracts tokens such as overall topology, outline, main lines, and main textures in the selected interval. Then, through the chained associative activation process, the machine obtains relevant memories in the memory network and combines these memories into expected models of different weights according to their weights.
    
The machine uses the decision-making process to decide whether to further recognize the input information or respond to the input information according to the activated reward and punishment tokens (the magnitude of the activation value of the reward and punishment tokens is the expected reward and punishment value).
	If the machine decides to further recognize the input information, the machine further extracts "expected tokens" from the input information by imitating the relevant experience of obtaining "expected tokens" in the past. Therefore, the machine continuously extracts tokens from the input information through the attention mechanism through iterations. In each extraction process, different sensors may be used, for different recognition intervals, and different resolutions may be adopted. Therefore, for the same input thing, the machine may extract tokens of different types, different intervals, and different resolutions, and use these token combinations to form a "hierarchical representation" of the same thing. "Hierarchical representation" refers to extracting tokens of information sequentially in a way that gives priority to overall features at low resolutions within an interval.
	Expected models are constructed using high-activation value tokens. The theoretical basis for this is that these high-activation value tokens come from two parts: First, the common features of similar things; because common features widely exist in similar things, they have high repetitiveness, so they are usually tokens with high memory values. Therefore, in our scheme, the machine's method of recognizing information is to first recognize large categories through common features (obtain abstract concepts), and then gradually add more tokens through iterative methods to limit the scope (from abstract concepts to specific concepts).
    
	Another source of high activation values is: There are tokens similar to specific memories in the input tokens. These specific tokens will directly activate the tokens in memory due to similarity activation, and other tokens with high memory values that have a proximity relationship with them are also likely to obtain higher activation values. Due to the short activation path, in the relationship network, special tokens will activate specific "expected models". This is a way to quickly locate expected models through special tokens.
	So the recognition process of input information is to identify which large category it belongs to through common features, and then determine which specific subclass it belongs to through unique features. The machine continuously iterates and adds tokens for recognition through selective attention. In this process, the activation values of previously activated tokens will fade over time. If they are reactivated by newly input tokens, their activation values will remain. If they are not related to the newly input tokens, their activation values will gradually fade and gradually withdraw from the decision-making process.
    
	Therefore, the "world model" contains two layers of meaning: 1. The machine actively iterates to recognize information in a "pattern recognition" way. And the pattern refers to the "world model." 2. The machine autonomously selects the content that needs to be recognized according to the magnitude of "value of advantages and disadvantages." Therefore, the "world model" is established around "value of advantages and disadvantages." Information that has nothing to do with "value of advantages and disadvantages" will be ignored by the machine.

    \subsection{We have achieved logical reasoning ability compatible with neural networks}
    Professor Yushua Bengio, one of the three giants of deep learning and a Turing Award winner, believes that the most important step to achieve general artificial intelligence is to combine neural networks with causal reasoning. In fact, our scheme has already achieved this: the memory network is a fully connected neural network. From input tokens to the activated "world model" is causal reasoning about the organizational way of the objective world; from input tokens to the activated "combination of subjective tokens" (tokens representing needs, emotions, rewards and punishments) is causal reasoning between the objective world and the machine's own needs. Therefore, we have achieved "combining neural networks with causal reasoning."
    
    \subsection{We have achieved hierarchical "general decision-making ability."}
The machine only reinforces learn one task: "How to meet its own needs?" and only processes one task "How to meet its own needs?" Therefore, for our machine, decision-making is "facing its own needs," while for other current AI schemes, decision-making is facing various "tasks themselves."

	Information input generates various associations, some good and some bad. Reducing the probability of occurrence of tokens that bring "punishment" and increasing the probability of occurrence of tokens that bring "reward" is "general decision-making." This is similar to human decision-making, so it is general. With the pre-positioning of the reward function, the machine has "decision-making ability" for intermediate steps. With the general goal of "seeking advantages and avoiding disadvantages," the machine can achieve "general decision-making" ability.
    
	In any environment, the input information of the machine includes all sensor information. So at any moment, the environmental information where the machine is located is part of the input information. The machine's interactive decision-making with the environment includes two aspects:
    
1, Find the optimal decision.

2, Execute the optimal decision.

	But these two steps are not separated. They are intertwined and processed in parallel. The first problem that "general decision-making" needs to solve is: What is the reward function? In GPT-4 and AlphaGo, the reward comes from external feedback. In our AGI, the reward comes from the "reward" and "punishment" symbols activated by external information, and the magnitude is their activation values.

\subsubsection{Step 1: What is the purpose of machine decision-making?}
	When information input (external + machine's own monitoring information) is input, some reward symbols and punishment symbols are activated.
    
	Each activation value transfer path from input to reward symbols and punishment symbols is a potential logical link that generates rewards or punishments.
	If each underlying feature on this logical link is truly realized, then the reward or punishment propagated by this logical link is also realized.
    
	So the machine's response to any input information is the same: increase the probability of occurrence of reward logic chains and reduce the probability of occurrence of punishment logic chains to achieve the purpose of seeking advantages and avoiding disadvantages.

\subsubsection{Step 2:With the top-level purpose, how to execute?}
How to increase the reward link and reduce the occurrence probability of the punishment link?

	That is to increase or reduce the realization probability of the combination of high-activation value tokens on the link. The combination of high-activation value tokens on the link is the high-weight token combination of this link. When they are true, the activation value propagated along this link is true, so the finally activated reward or punishment is also true.
    
How to operate specifically?

	On the activation value transfer path from input information to reward and punishment symbols, select the N tokens with the highest activation values. They are the top-level implementation paths that lead to rewards or bring punishments to be true. The goal of the machine is to make the tokens on the reward path realized (that is, imitate past experience and make them appear in the input information); make the tokens on the punishment path not realized (that is, imitate past experience and avoid them from appearing in the input information). Therefore, on the logical path from input to reward and punishment, selecting the N tokens with the highest activation values and including their activation value propagation paths is the top-level implementation path.
    
	Why does the machine only select the N tokens with the highest activation values? Because these tokens are either because they are representative tokens of a certain type of thing, so they have high memory values and thus obtain higher activation values; or they are tokens closely related to the input information. Due to the small number, it is equivalent to having few attribute restrictions, so the concepts most closely related to them are usually "abstract concepts."
    
Since language symbols are used very frequently, language tokens often obtain high activation values and become the core tokens with the highest activation values in the combination of "abstract concept" tokens, making language symbols representatives of concepts themselves.

So the process of the machine establishing decisions is to prioritize "abstract concepts" and then gradually add more tokens to form more specific concept combinations. This is a top-down, step-by-step decision-making and execution process. We call this process "segmented imitation."
	A specific example of the method of segmented imitation:
    
	Assume that the set of input tokens is taken as A and the set of response tokens is taken as B. Through the chained associative activation process of A and B, the machine looks for those high-activation value tokens. These tokens are tokens that have a connection relationship with both A and B. Because they obtain activation values from both A and B, they become high-activation value tokens. They are the intermediate bridge tokens connecting A and B. This process can be iterated to achieve top-down, layer-by-layer decision-making.
	Why only select the N tokens with the highest activation values at each expansion? This is because past experience and current reality cannot be completely matched. By only using tokens with the highest activation values, it means that the "model" composed of them is either abstract (with a wide range of application) or closely related to input tokens (with good matching). The purpose of only selecting the N tokens with the highest activation values is to achieve experience generalization. Therefore, in our scheme, experience generalization is automatically achieved by selecting high-activation value tokens and discarding low-activation value tokens, thereby expanding the scope of application of concepts. Of course, expanding the scope of application of concepts means using more abstract concepts.
    
	For example, the machine has the experience of using a hammer to hit a nail. In the case of needing to hit a nail and there is no hammer, in order to achieve the first-level goal (complete the task, obtain a reward, or avoid being punished), the token combination representing a hammer may be included in the activated logical link. Then, these token combinations become second-level goals.
    
	According to the chained associative activation process in the memory bank, the machine may find that the tokens representing "hard objects" obtain the highest activation value. The tokens representing "hard objects" are only one token in the concept of a hammer. And the concept of a hammer is a set of tokens that often appear with a hammer. They are a local close relationship network formed in the memory bank due to adjacency relationship and high memory value through the chained associative activation process + memory and forgetting mechanism. This relationship refers to the activation value transfer relationship. If contour tokens, language tokens, or symbol tokens related to hammers appear in the input, then undoubtedly the tokens with the highest activation value in the "hammer concept" may be hammer contour tokens or tactile tokens of using a hammer. But if there are no contour tokens, language tokens, or symbol tokens related to hammers appearing in the input, then the tokens with the highest activation value at this time may be the tokens representing "hard objects." If tokens related to stones appear in the input, then tokens related to both hammers and stones (such as weight feeling, size, hardness feeling, etc.) may obtain higher cumulative activation values and be selected as the top N high-activation values. So the "hard object" token is a common token of hammers and stones and also an experience generalization bridge between the two. When the machine selects a decision-making path, it will first select the experience related to "hard objects" with the highest activation value. Then lower the activation value selection criteria and select more tokens to enter the "world model". At this time, the "hard object" concept will be limited to the "stone" concept because more tokens are added. Therefore, it can be seen that in 	our scheme, the experience generalization process is automatically completed.
    
Some paths may bring both rewards and punishments, which makes it difficult for the machine to converge when counting reward and punishment values. If the machine finds that there are multiple conditional probabilities in the reward and punishment value statistics and there is no definite indication. At this time, under the principle of seeking advantages and avoiding disadvantages, the decision is to further determine the truth of each condition.

	Therefore, in our scheme, the machine's decision-making is hierarchical. In a decision-making path, a large number of local decision-making and execution processes may be nested. But at any time, the machine's only goal is "seeking advantages and avoiding disadvantages." All decisions are derived from this goal. Therefore, the machine's decision-making is also very flexible. It changes according to the environmental state at all times and there is no preset process. The only preset process is: "seeking advantages and avoiding disadvantages."
    
	The above process is iterated, and new reward and punishment symbols are activated each time. The machine counts the activation values of these reward and punishment symbols until it finds that the activation values of the reward and punishment symbols converge. At this time, the machine has established the optimal response path.
    
	The machine's decision may be a response to input information, or it may be to find more information to judge the truth of the conditions in conditional probabilities, so as to better "seek advantages and avoid disadvantages." In either case, the machine increases or reduces the probability of occurrence of specific tokens by imitating past experience. At any time, after new information is input, the new information will update the activation value distribution in the memory bank through the chained associative activation process. At this time, the machine needs to re-count the reward and punishment information and find the optimal decision again according to the new state. This process is ongoing all the time.

\subsubsection{Step 3: With a plan, how to execute?}
Execution is to increase or reduce the probability of occurrence of specific tokens by imitating past experience.

Select a small number of highest activation value underlying features → abstract decision-making path.

Add more high-activation value underlying features → concretize the abstract decision-making path.

Steps 1 and 2 above are iterated until the decision is decomposed into executable drive commands. Drive commands: send waveforms to speakers, send drive commands to motors, send display data to displays, send setting parameters to expression display systems, etc.

New input information may be encountered at any time. New input information will change the activation values in the memory bank and change the reward and punishment situation. Therefore, during the implementation of the optimal. 

\subsubsection{We have solved the following problems}
(1) The problem of how to "establish common sense."

	In our scheme, the essence of "knowledge" is the common arrangement relationship of tokens in time and space, and the prediction of the potential advantages and disadvantages for the agent by different token arrangements. The arrangement relationship of tokens in time and space is essentially "causality." These arrangement relationships of tokens in time and space are not simply temporal and spatial proximity relationships, but relationships that the agent summarizes and can repeat. The temporal and spatial spans they actually span may be very large. However, through the chained associative activation process, these tokens with large temporal and spatial spans form a close activation value transfer relationship, which is knowledge. If knowledge contains tokens related to "needs," "emotions," and "advantages and disadvantages," this can predict potential advantages and disadvantages. Therefore, the arrangement of tokens represents "knowledge." And those common arrangements are "common sense."
(2) The problem of "whether machines can have consciousness."

	We have solved how to endow machines with "self-needs." Therefore, the machine can make autonomous decisions, self-evolve, have its own emotions, and pursue "self-needs." So our machine is "conscious."
    
(3) The "general decision-making" problem.

	The machine makes decisions according to "seeking advantages and avoiding disadvantages" for any task. The tasks given by humans are by-products of the machine's pursuit of "self-needs."
    
This is the same as when you complete the tasks assigned by your boss. You also complete the tasks assigned by your boss in the process of pursuing "self-needs." If there is a conflict between the two, you will also make various flexible decisions according to seeking advantages and avoiding disadvantages, test the real intentions of your boss, and consider the bottom line of your boss. So your decision-making will be very flexible.

(4) The "language understanding" problem.

	Because we have not destroyed the original temporal and spatial relationships of tokens. The temporal and spatial sequence of tokens represented by language sequences can be understood and imitated. Therefore, machines can directly learn various skills through language like humans. After reading the oven manual once, you can start baking bread \cite{ref5} \cite{ref6} \cite{ref7} 

\subsubsection{Advantages of our scheme}
Therefore, we believe that our path is a feasible path to AGI. Its advantages lie in:

	(1) It can handle tasks that cannot be trial-and-errored in large numbers.
    
For example, autonomous driving, domestic nanny, taking care of the elderly, accompanying children, and engaging in "workers, peasants, soldiers, scholars, and businessmen." This is because we are "human-like" AI and can directly obtain human existing experience through language without having to summarize every skill through a large amount of training.

	(2) It can achieve general decision-making.
    
Because in our scheme, the machine only learns one thing "how to meet self-needs" and only processes one thing "how to meet self-needs." So it has general decision-making ability.

	(3) It can solve the "hallucination" problem.
    
Large models only have "common phrases" obtained through local statistics and do not retain the input information itself. That is, there is no memory of facts. In our scheme, we first store memories and then extract common information from memories. So we come with a "fact database," and facts will be output as the most matching results.

(4) Safer.

	At present, artificial intelligence has a single goal. In terms of decision-making, it is an artificial intelligence that "achieves the goal by any means." Such artificial intelligence does not have values similar to humans but only focuses on the reward function given externally. So it will not consider anything other than the reward function. Moreover, the machine's decision-making is still a black box, and humans do not know its thinking process. If such artificial intelligence fully controls human life, it is entirely possible that due to misinterpretation, it will bring incalculable disasters to humans.
	In our scheme, the "type of needs" of the machine can be preset, values can be trained, and can be aligned with human values. Facing any task, the machine's decision-making is to count all reward and punishment information, including the reward and punishment information brought by values. So the machine will not exhibit "extreme" behavior. Moreover, in our scheme, decision-making is visible, modifiable, and a "white box." Therefore, our scheme has higher safety.

\subsection{A simple implementation example}
\subsubsection{Creating the Memory Bank}
Since this project is a personal research project lacking funds and computing power, we adopted a very small scale to attempt to verify this idea. We only used a small amount of text data and MNIST (handwritten digits) data as training data.

The first step is to read in each word in the text data, then convert each word into a vector and store it in the memory bank. As shown in the following code, the first column represents the storage time. Here, we don't need to record the absolute time but use increasing integers to represent the time sequence of storing in the memory bank. For example, when storing the first token, its id is 1; when storing the second token, its id is 2. This facilitates quick queries.

The second column represents the stored Token itself. It is a mixed vector containing multiple dimensions, such as any one of text, graphics, voice, or values obtained from different sensors (such as temperature, pressure, smell, taste, etc., and each type of sensor data constitutes a dimension). However, each Token only has data in one dimension. These data represent the data obtained by the machine using eyes/ears/nose/mouth/skin, or even infrared/ultraviolet rays, or any other sensors. For example, it may be graphic Token data (any feature point extracted from a single MNIST data); it may also be voice Token data (a syllable slice obtained from voice); it may also be real-time data transmitted by sensors such as temperature and pressure. For example, the second column can contain multiple sub-fields, which are as follows:

The first sub-field is called Vision (4 channels, a 28*28 vector);

The second sub-field is called Auditory (a 2-component vector for the left and right ears);

The third sub-field is called Tactile (a 4-component vector, including the magnitude/direction of the perception force + temperature perception);

The fourth sub-field is called Taste (an 8-component vector for 8 types of taste);

The fifth sub-field is called Smell (an 8-component vector for 8 types of smell);

The sixth sub-field is called Reward (an 8-component vector, dividing human rewards into 8 different types);

The seventh sub-field is called Punishment (an 8-component vector, dividing human punishments into 8 different types);

The eighth sub-field is called Emotion (dividing human emotions into 8 different types).

In our implementation example, we only adopted 4 dimensions including text, graphics, rewards, and punishments.

The third column is the memory value, representing the pre-training weight of the Token, which is a double-precision number;

The fourth column is the activation value, representing the inference weight of the Token, which is also a double-precision number.

\subsubsection{The First Stage of Training: Simple Repetitive Statistics}

In our implementation example, for the sake of simplicity, we just read in each word in the text and took each word as a token, then turned it into a database record. In the initial stage of training (about ten thousand words), we simply assigned a memory value (mv) of 1 and an activation value (av) of 1 to all records. Thus, an initial memory bank was formed. Then, every time we read in a word, we would search for the same word in the memory bank. In the original memory bank, each time a word was found, its repetitiveness was considered to have increased by 1, so we would update the corresponding record. The way to update was to increase its memory value by 1. Then, the newly read word was still stored in the latest position of the memory bank in the order of time with mv set to 1. Through this way, we trained text with about one hundred thousand words.

The memory value (mv) of each record in the database represents the repetitiveness of this word. However, the repetitiveness of various word combinations has not been reflected. In our scheme, the core is to find the repetitiveness of token combinations and assign higher weights to those combinations with high repetitiveness. So, we need the next stage of training.

\subsubsection{The Second Stage of Training: Association Statistics}
\paragraph{Simple Association Statistics}

This time, we continued to read in the input Tokens. Every time we read in a Token, we would search for the same (or similar, such as graphics) Tokens in the memory bank. For each found Token, its memory value was increased by A, and for the 2N adjacent records (such as the N records before them and the N records after them), their memory values were also increased by A1, A2... An respectively. In our example, A was set to 1, N was set to 1, and A1 was set to 1/4 * A. After completion, we continued to put the new Token into the last position of the memory bank with mv = 1 and av = 1.

\paragraph{Presetting Simple Knowledge}
To preset simple knowledge, you just need to write it into the memory bank according to the format of the database and the time sequence in which these tokens were obtained by sensors, and adjust their memory values according to human experience and their repetitiveness.

For example, for several basic graphic tokens that can represent a smiling face, each Token is a record. Then, according to the principle of prioritizing overall feature tokens, they are stored in the memory bank and assigned a high memory value. Then, tokens with the second highest representativeness are stored in the memory bank and assigned a relatively high memory value. Then, a special symbol is used to represent the reward-related Token. This special symbol can be a separate component of the second column vector; it can be a specific symbol or even a code. Then, a large memory value is assigned to it to form a record and place it after the smiling face Tokens (so that the id of the reward Token is close to that of the tokens related to the smiling face).

In our example, we defined the token column as a 16-dimensional vector. These 16 dimensions are respectively: label (such as the word itself), Vision (such as the graphic of the word text, or MINIST handwritten digits), Auditory (such as syllable data), Reward and Punishment, Perception 1 (such as temperature), Perception 2 (such as pressure and tactile perception), Perception 3 (such as gravity perception).... Perception 12 (to be determined). Obviously, each dimension can also be a multi-dimensional vector (such as Vision).

In our implementation example, we only trained a small amount of text and handwritten digits (MNIST library), so for text tokens, only the label column has data, while for handwritten digits, only the label column or the Vision column has data. In our example, we assigned a positive value of 1 in the reward dimension (representing the default reward size. For example, it can also be set to -2, representing a large punishment). Then, we preset a value for the mv value of this record to represent the repeated probability of the appearance of rewards or punishments after the relevant Tokens appear in the input.

The preset reward or punishment knowledge will be adjusted through subsequent learning to form new reward or punishment knowledge and thus adjust the behavior of the machine. So, very little and simple innate knowledge is required. However, there are two types that are essential: one is the simplest way of communication with humans, such as allowing the machine to receive a reward when it sees a smiling face. As for the more meanings of the human smiling face, they will be gradually enriched by establishing new connection relationships through subsequent learning. The second is about the safety of the machine, such as giving a reminder when the power is low; or avoiding a cliff when seeing it, etc., just the most basic knowledge is enough.

\paragraph{Chain Association Activation}
This time, every time we read in a Token, we turn it into a temporary record. Its memory value is a preset initial value mv0, and its activation value is a preset initial value av0.

In our example, after the input tokens obtain the initial activation value av0, they will transfer the activation value to similar Tokens in the memory bank. This is the similarity activation process.

The similarity activation process is to search for similar Tokens in the memory bank and then spread the activation value according to the activation value transfer function. In the activation value transfer function av-added = tf(similar, mv, av0), av-added is positively correlated with the similarity of the token comparison (such as the comparison of two graphic Tokens), positively correlated with the mv value of the token being transferred in the memory bank, and of course, positively correlated with the activation source av0 that initiates the association activation.

In addition, since we adopted the method of accumulating repetitiveness by +1 to count the repetitiveness of a single word in the early stage, the mv value is usually large. Therefore, we performed nonlinear compression on it in the activation value transfer function tf, for example, using the Logistic function to map the x-axis from the interval of 0 to positive infinity to the interval of 0 to 1.

Each Token activated by similarity, if its activation value is greater than the preset threshold, will also initiate the adjacent activation process. The adjacent activation process is to transfer the activation value to adjacent Tokens; the closer the time position, the greater the transfer coefficient; the higher the memory value, the greater the transfer coefficient. Of course, in order to avoid repeated transfers between two Tokens, each transfer process is unidirectional. In our example, adjacent means the records with adjacent ids (such as the records with id + 1 and id - 1). So if the record in the id row obtains an activation value exceeding the preset threshold, it will transfer the activation value to the id + 1 row. And the propagation rule is av-added-adj = tf(mv, av0, k1), where mv is the mv value of the token being transferred, av0 is the activation value of the token that initiates the propagation process, and k1 is the adjacent activation propagation coefficient between Tokens with an id distance of 1. av-added-adj is positively correlated with mv, av0, and k1. Similarly, if the updated activation value of the id + 1 row exceeds the preset threshold, it will transfer the activation value to the id + 2 row according to the same rule. Similarly, if the updated activation value of the id + 2 row exceeds the preset threshold, it will transfer the activation value to the id + 3 row according to the same rule. This process continues until the updated activation value is below the preset threshold. Obviously, the upper limit of the tf function is 1. Similarly, if the record in the id row obtains an activation value exceeding the preset threshold, it will transfer the activation value to the id - 1 row and transfer the activation value in the direction of decreasing id according to the same rule.

In the actual training process, we found that many punctuation marks and common conjunctions appear very frequently in the memory bank, and they will start an excessive activation value propagation process from the input. So for these words, we need to establish a Stop tokens list to reduce their initial activation value av0, thereby suppressing their excessive activation value propagation.

When the similarity activation and adjacent activation for a single input are completed, the program enters the chain activation process. In the chain activation process, the program preferentially selects the top N combinations with the highest activation values. So, we need to first define the selection criteria for the top N combinations with the highest activation values. We look for the combination with the highest sum of activation values within a certain range of records. For example, taking consecutive M (such as 100) tokens as a group, if their cumulative sum of activation values is the highest, this group of Tokens will be taken as the combination with the highest activation value. Then, this group of Tokens will be taken as the expected model. It is the expected concept activated by the machine under the current input.

To simplify the calculation amount, in the actual implementation process, we first find the Tokens with the highest activation values, then select the top 100 as the center, and slide the statistical window forward and backward to look for the expected model. Then, we select the top A (such as 20) of the expected model with the highest cumulative activation values as the optimal expected model.

Then, among these A combinations, we look for the Tokens that have the highest activation values among the top X Tokens that did not appear in the input. Then, taking the activation values they obtained as the initial activation values, we initiate their similarity activation process and adjacent activation process again. This is association activation.

After the association activation is completed, the machine again looks for the combination with the highest sum of activation values within a certain range of records and again adopts a process similar to the above to look for the Tokens among the top X Tokens that have not initiated the association activation process and initiate the association activation again.

This process can be iterated with a depth of D. In our example, D is 2.

As long as the Tokens are activated, it means that they have a direct or indirect association with the current input. It represents that under the current input excitation, these Token combinations have higher weights. This is equivalent to the way that in LLM, the attention mechanism is used to readjust the weights according to the input information. The difference is that we have retained the original organizational form of the tokens. That is to say, we have retained the factual memory.

In our scheme, the weight matrix is distributed to each Token, and the memory value is used to represent it. And the activation value represents the current weight redistribution.

\paragraph{Memory Value Update}

After a single input is completed, the machine updates their memory values according to the activation values obtained by the Tokens in the memory bank according to certain rules. The update function is: mv-delta = tf(mv, av), where mv-delta is the increment of the memory value; mv is the original memory value of the Token, and av is the activation value currently obtained by the Token.

In our implementation scheme, forgetting is crucial. The machine needs to forget to obtain the combination ways and relative weights of common Token combinations. We initially adopted the method of subtracting a delta from the entire memory bank every once in a while (for example, when the id increased by 100K). However, we found that this method was not effective. Due to the large amount of data, this linear decrease did not conform to the actual human memory process, and it was easy for a large number of tokens to be forgotten.

Therefore, in the forgetting process, a nonlinear function needs to be used to imitate the human forgetting process. One method is to compress mv to the range of 0 to 1 in the activation value propagation function tf, and let the central value corresponding to 0.5 increase as the training time (the largest id) increases. So, the Tokens with small mv values will gradually have their weights decrease as time increases, which imitates the forgetting process.
To accelerate the calculation, it is also possible to directly compress mv to the range of 0 to 1 and let the central value corresponding to 0.5 increase as the training time (the largest id) increases. Then, set a memory value threshold, and records below this threshold will be deleted (completely forgotten).

\paragraph{The Third Stage of Training: Machine Autonomous Behavior}
The entire chain association activation process + memory and forgetting mechanism essentially imitates the human memory process to discover common combinations of Tokens and can retain the original organizational form of Tokens.

Then, through chain association activation, based on the current input Tokens, based on the statistical weights (the memory values of Tokens and their organizational ways), the weights (activation values) are redistributed. This is essentially the same as the attention mechanism in Transformer.
These activated Tokens, the organizational ways of these Tokens, and their activation values constitute the latent space corresponding to the input Tokens information. The latent space here is similar to the latent space formed by LLM through the attention mechanism + Encoder. The machine can generate output by using the decoder through the latent space. If a decoder is trained through self-supervised deep learning in a similar way, then our scheme is similar to the current LLM. The difference is that another method of implementing the attention mechanism is adopted and the original training materials are retained. In this case, in order to enable the machine to have decision-making ability, reinforcement learning is still required. This goes against our original intention of creating human-like intelligence.

Therefore, we have created a new general decision-making scheme. The core idea of this general decision-making scheme is to let the machine have innate needs and then let it learn how to meet its own needs through learning. The current technical scheme is to do reinforcement learning for specific tasks. Such a technical scheme can only train robots for limited tasks in limited scenarios and cannot achieve universality. So, we first create a "machine baby" and let it learn how to survive and grow in life.

So we need to be able to preset innate knowledge for the machine, and this innate knowledge should be integrated with acquired knowledge and can be called by each other seamlessly. So it cannot be external rules; it must be part of the machine's knowledge network. This innate knowledge includes basic reward and punishment information, basic communication ability with the outside world, such as expressing "hunger", "crying" when uncomfortable, and being able to recognize the simplest "praise" and "punishment". With preset knowledge, the machine has its own needs, rewards and punishments, emotions, and autonomous behaviors. We will use our example to illustrate how the machine creates autonomous behaviors.

In our example, the first part of the innate knowledge preset for the machine is: the innate communication means between humans and the machine.
For example, when humans input a simple graphic like a smiling face ICON to the machine, the machine can recognize that it is a reward.
Specific implementation process: We can input a large number of various smiling face ICONs. For each smiling face ICON, extract the overall features according to the principle of prioritizing overall features and take them as a Token to form a record and store it in the memory bank. Then, extract the local features, and take each local feature as a Token to form a record and store it in the memory bank. These records are stored in sequence according to time, so their time labels (ids) are adjacent.

Through simple repetitive statistics, then through simple association statistics, and then through chain association activation, as well as memory and forgetting mechanisms, the machine obtains a memory bank. In this memory bank, the order of Tokens is the order in which they are stored, but the memory value of each token is related to the repeated frequency of their appearance in these smiling face ICONs used as training materials. Those Tokens with a high appearance probability will obtain a higher memory value, that is, they have a higher weight.

This local network is physically just a segment of records. But in terms of logical composition, the Tokens within it are network nodes, and the network connections are their activation value transfer relationships. And the activation value transfer relationship is determined by their mutual similarity, storage position (id), and the size of the memory value. Those Tokens with relatively large propagation coefficients between each other form a "concept". Obviously, a "concept" doesn't have a clear boundary. Under different input stimuli, the activation values obtained by the Tokens inside it are not the same, and the distribution of these activation values represents the weight redistribution of the "concept" under the current input stimuli.

Then, we insert reward Tokens after this segment of memory. In our example, we only need to insert a special record. This special record has a positive value of 1 in the reward dimension of the Token (representing the size of the reward). Then, we preset a relatively large value for the mv value of this record, representing that there is a highly repetitive time relationship between the smiling face ICON and the reward.

In this way, when a smiling face ICON appears in the input, the machine also extracts the overall features in the way of prioritizing the overall features, takes them as the first batch of input Tokens, assigns initial activation values to these Tokens, and then spreads these activation values in the memory bank through the similarity activation process. Due to the similarity of the Tokens, the Tokens in the memory bank preset by us regarding the smiling face ICON may obtain activation values. These activation values will spread the activation values to the reward Tokens through the chain association activation process, thus activating the reward. And the size of the activation value of the reward Tokens represents the size of the reward. In this way, the machine can recognize that the input is a reward.

Similarly, we can use the same process to preset a communication method for the machine to express punishment. For example, when the outside world needs to express punishment to the machine, it inputs words like "Your behavior like this is not good." To achieve this, we only need to train similar expressions to form Token combinations with certain weights, insert records representing punishment Tokens in the adjacent positions, and assign relatively high memory values to these records. In this way, after such words are input, the punishment Tokens will be activated, and the machine can recognize that the input is a punishment.

Similarly, the machine also needs to establish the innate knowledge to express itself to humans. For example, when the cumulative activation values of all the punishment Tokens reach a relatively high threshold, the machine will express a "depressed" expression to the outside world according to the preset program; when the cumulative activation values of all the reward Tokens reach a relatively high threshold, the machine will express a "happy" expression to the outside world according to the preset program; another example is that when the machine's power is insufficient, the machine will express a "hungry" expression to the outside world according to the preset program, or directly display "I'm hungry" on the display screen.

In our example, the second part of the innate knowledge preset for the machine is: the machine's innate needs.

Firstly, the machine must have the innate need to survive. For example, through the internal monitoring program, it monitors the power level of the machine in real time and converts the machine's power consumption into a "power bill". Every time the machine outputs a response, it will consume a certain amount of "power". When the machine's power is too low and falls below the "warning line", it will trigger Punishment A; when the machine's power falls below the "critical line", it will trigger Punishment B. When the power value falls below the "warning line", the monitoring program will assign activation values to the Tokens representing insufficient power in the "innate knowledge", that is, the innate memory. And in our preset memory, behind the Tokens representing insufficient power are some punishment Tokens with relatively high memory values. So insufficient power will activate the punishment Tokens, and the machine will recognize the punishment. Similarly, the machine will monitor its own power consumption in real time. If it finds that its power bill is too large, it will activate the Tokens related to "high power bill" in the innate memory. Usually, punishment Tokens are adjacent to these Tokens. So the machine will recognize that "high power bill" is a type of punishment.

In our example, we are unable to monitor power and power bills in real time, so we give the machine a virtual "power" value of 100. After the machine outputs a result each time, we subtract a delta from the "energy" value. Through this simulation process, after the machine outputs many times, the power will drop below the "warning line", resulting in punishment.

The purpose of doing this is to make the machine as "lazy" as humans. And "laziness" is the original driving force for humans to spontaneously create tools, improve efficiency, and promote the progress of science and technology.

Secondly, we need to preset correct values for the machine, such as the pursuit of obtaining rewards. These can be achieved by making the machine pursue the maximization of rewards in the decision-making process.

In our example, every time the machine reads in a handwritten digit, it can choose to output this digit or not. If it chooses not to output it, the response from the outside world is to input "Your behavior like this is not good." This actually corresponds to a punishment. After several attempts, the machine will learn that choosing not to respond will bring punishment, so it will give more responses. When the number of responses is large, the machine will receive a greater punishment due to the power warning. So it needs to weigh the pros and cons and choose whether to output or not. If by chance the digit output by the machine is "7", we will input a smiling face ICON to the machine to represent a reward. The machine will receive the reward and get a power supplement to 80. Since the power has changed from the warning state to 80, the machine is free from punishment, and all these will be saved as memories. After repeated times, the machine will establish a closer relationship between the digit "7" and the "increase in reward and punishment values" (because the Tokens representing these actions are close in time).

After a period of training, the machine will learn to reduce the output of other digits because it will consume energy, and learn to increase the output of the digit "7" in order to obtain rewards.

This seems to be a general learning behavior. Because any behavior of the machine can be trained through rewards, and the specific implementation process is explored by the machine itself. Although this method seems to be a kind of reinforcement learning, it cannot achieve "human-like learning ability". In fact, we found that by taking the activated latent space records, including all the latent space records (including id, Tokens, memory values, and activation values) screened by the expected model as the actual input and using a decoder, the machine can still generalize the time series decision-making process. We think this is because our knowledge network retains the original organizational information of the Tokens. The machine can obtain human's existing experience through language, which is equivalent to the machine having gone through all the trial-and-error processes that humans have gone through. So it can draw on human experience to make correct decisions. It can select the response path with the maximum benefit from the different results output by the decoder as the decision. Then it can take the response path as a virtual input and evaluate the response path with the maximum benefit again until the maximum value of the evaluation converges.

Our original intention was to make the machine count the activated reward values, then select the path with the maximum reward, and then use the segmented imitation method to increase the realization probability of the Tokens on the path leading to the reward and suppress the realization probability of the Tokens on the path leading to the punishment, so as to achieve true human-like intelligence. However, we have limited resources (this project is a personal amateur research project without the support of funds and computing power), and we are unable to achieve the above goals.

However, we believe that "human-like intelligence" has two core characteristics: one is having "self-needs", so that it can create tasks by itself. The other is being able to directly obtain others' experience through language, so it doesn't need to do "reinforcement learning" for everything. The above two characteristics are exactly the differences between our scheme and the current neural network + reinforcement learning technical route. And for characteristic one, it can be achieved by presetting the innate memory network representing needs and communication. For characteristic two, during the training process, language Tokens are always input together with other Tokens represented by language symbols. 

This is just like how we teach babies language in life. So there is a neighboring activation value relationship between language Tokens and the Tokens they represent, and this relationship is constantly reproduced in life and continuously strengthened in the memory bank. This is the process of mother tongue learning. Then, the machine can use language information alone to obtain the multi-modal information represented by the language information. Because after the language information is input, not only the language Tokens are activated, but all the related multi-modal Tokens are activated. The combination ways of those high-activation-value Tokens in time and space represent the development process of a certain kind of things. This is the cognition of the universal laws of external things, which we call "objective common sense". And in this activation process, the activation values of the activated reward and punishment Tokens represent the potential "reward and punishment" of this kind of process to the machine itself. This is the machine's cognition of the impact of the development of external things on itself, which we call "subjective common sense". After the machine has "objective common sense" and "subjective common sense", the machine has "common sense". And common sense is the machine's cognitive system and value system. So the machine can imitate these high-activation-value Token combinations to obtain rewards or avoid punishments. This is "human intelligence". It is precisely because we have retained the time and space organizational form of the Tokens themselves and extracted the universal laws from them that we can use the universal laws to imitate and solve specific problems. In the existing neural network + reinforcement learning technical route, the time and space organizational form of the Tokens themselves is not retained, so the laws it discovers and the laws of human cognition are written in different symbolic languages. When dealing with language, the translation network between machine laws and human laws can be achieved through a large number of corresponding materials. But when dealing with multi-modal and involving real-time interaction, this translation process is difficult to complete. And reinforcement learning itself is essentially a neural network training process involving time interaction. Reinforcement learning is such a neural network: all inputs and intermediate processes are taken as input data, and the final reward is taken as output. The machine adjusts the parameters of the neural network through the trial-and-error method, and it is essentially no different from a multi-layer neural network. However, since there are a large number of time processes in our world that cannot be tried and errored (such as taking care of babies, building buildings), the way to achieve AGI must be that the machine needs to draw on all the laws accumulated in the history of human civilization, that is, to learn laws from humans through language. And these laws are essentially the permutations and combinations of representative Tokens in time and space. This is the fundamentally different starting point between our scheme and the current AI technical route.

In our simple example, we first train the text because the source of text materials is extensive and the training is simple. Then, on the basis of completing the text training, we continue to train the handwritten digits. Since the content of the subsequent training is continuously added to the memory bank, it will not damage the previous training results and the preset innate knowledge. So our training will not damage the existing abilities.

When training handwritten digits, by calling different feature extraction functions in OpenCV, we take these features as graphic Tokens simply according to the principle of prioritizing overall features and then local features, and form a record and store it in the memory bank. In this way, the entire model training process is completed.

When using the model, the input Tokens will undergo the chain association activation process, and at the same time, they will be stored in the memory bank as a record. And after each input is completed, the memory bank will update the memory values according to the activation values. So the process of using is also the process of training. Our scheme is similar to training "humans", without a clear dividing line between training and using, and there will be no catastrophic forgetting of the original knowledge when learning new knowledge.

In our simple example, after the training is completed, every time a matrix data of a handwritten digit is input to the machine, the machine will recognize this digit by itself and randomly choose whether to give the label of the digit it has recognized. With the deepening of use, the machine will gradually find that the digit "7" can obtain rewards. When its energy is insufficient, it will tend to output the digit "7" more to obtain rewards and get energy supplements. This is its simplest active learning process.

Our example is very simple and is just used to demonstrate that our scheme can achieve: (1) Small-sample, cumulative learning without the problem of catastrophic forgetting. (2) It is a multi-modal training method and decision-making method with a single algorithm. (3) It can preset innate knowledge, including the machine's "self-needs", basic "reward and punishment" information, and the most basic "communication methods" with humans. (4) It can make the machine directly obtain human's existing experience through language, without having to do "reinforcement learning" for everything. And (3) and (4) are the unique abilities in our scheme, and these two points are the key steps to solve the problem of "human-like" robots. 

\section{A Workflow Diagram}

So the machine can iterates through the process shown in figure 2 to establish and execute decisions from top to bottom.
\begin{figure} 
    \centering
    \caption{A workflow diagram}
    \includegraphics[width=0.8\linewidth]{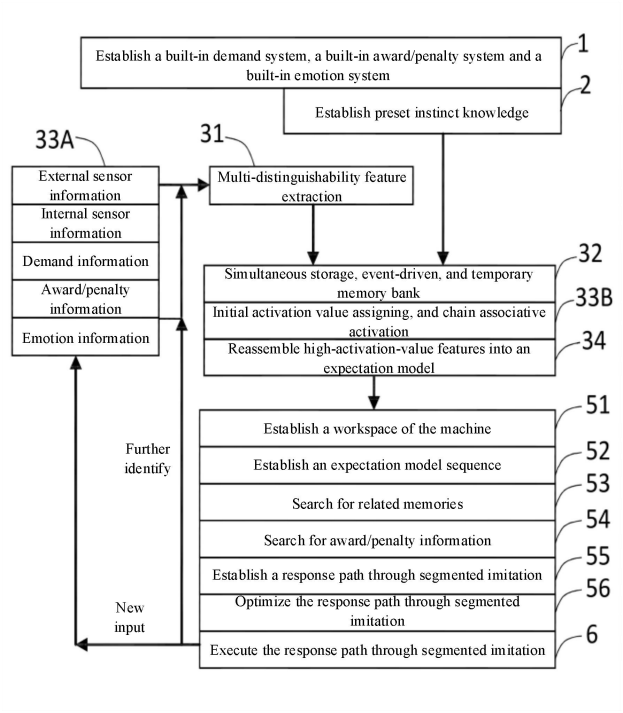}
\end{figure}

\section{Conclusion}

We believe that the development of artificial intelligence can be roughly divided into different stages: (1) The "feature exploration" stage. Before the advent of deep learning, it mainly focused on the "manual exploration" stage. After the emergence of deep learning, it centered on the "machine exploration" stage. (2) After the realization of the true attention mechanism (Transformer), as the "knowledge coordinate basis cluster" of machines and that of humans (concepts) were initially aligned, machines achieved "knowledge generalization". Facing human tasks, machines can demonstrate certain intelligence through "knowledge generalization".

The one-dimensional attention mechanism has brought about large language models. The two-dimensional attention mechanism has led to the generalization of images. The three-dimensional attention mechanism can achieve 3D creation capabilities. The four-dimensional (three-dimensional + time) attention mechanism can achieve the generalization of dynamic processes: it will bring about video generation and also robot services in limited scenarios.

However, we believe that only by adding the "vitality: the fifth dimension, self-needs" can the machine intelligence be truly endowed with a "soul". The reason is that if all the processes in the world and the results of these processes are presented to machines and then placed in a supercomputing center to form a neural network with hundreds of trillions or even quadrillions of parameters for training, AGI will be achieved, just like what the current LLM is doing. But we should note that language involves far fewer Tokens. One token represents an information unit, while the data in the real world is at least 3 to 5 orders of magnitude larger, and very little of it is electronic. The method of LLM is currently difficult to obtain the required data and also difficult to uniformly process such a huge amount of data. Therefore, data presentation for specific domains has been introduced: presenting all the processes within a limited range and the results of these processes to machines, which is reinforcement learning. So reinforcement learning + large models can ultimately only lead to AGI within a limited range. Since it is limited in scope, it cannot be called AGI.

These problems are difficult to solve: (1) Since the data is too large, a limited scope is adopted, which is deep learning + reinforcement learning. (2) It is difficult to obtain electronic data for many cases. For example, you can't really conduct experiments in real life to obtain data on whether a person is killed, unharmed, or disabled when hit by a car in various situations. So we can only adopt alternative ways: creating virtual scenarios, which is spatial intelligence. Electronic data under various behaviors are obtained in virtual spaces. But it will never fully match the real world. So, this is a wrong path.

Currently, text-to-video generation is very popular, but if machines don't have spatial knowledge or common sense of life, various bugs will always occur. This is training specific experts, and it is not the path leading to AGI. When AGI truly emerges, the real AGI will understand common sense of life, physical models, spatial transformations, and the correct time sequence among all things. It can read a novel and directly generate a movie that can be screened in theaters, and can also produce versions in various styles. By then, when looking back at today's text-to-video methods, we can only say that they are as primitive as tools in the Stone Age.

And the path taken by large models is doomed to be unable to achieve the "fifth dimension". However, our scheme can endow machines with "life", so it may become the real "artificial general intelligence". Our core ideas are as follows:

1. To achieve AGI, a human-like learning method needs to be realized.

2. Once the human-like learning method is realized, machines can directly obtain the experience that humans have summarized over tens of thousands of years through language, without having to try everything by themselves to know what to do.

3. To realize the human-like learning method, it is necessary to align the basis coordinate clusters of the machine information space and the human information space. Currently, LLM can indirectly align the basis coordinate clusters used in language for both through the attention mechanism. It is an indirect alignment because there is a translation process (transformation matrix) between the two basis spaces. Due to the large amount of data and the lack of electronic data of real-life processes, it is difficult for machines to establish the transformation matrix of the two basis spaces in the full-modal form. So it can only be limited to the local space to establish the transformation matrix of the local space, which is the ability that deep learning with attention mechanism + reinforcement learning can achieve. And we propose a new algorithm that enables machines to obtain the basis coordinate system in the full-modal form, which is directly aligned with the full-modal basis coordinate system used by humans. So the experience possessed by humans can be directly understood and executed by machines through language expression. This eliminates the need for reinforcement learning in every situation.

4. To achieve this, it is necessary to retain the original spatio-temporal organizational form of the full-modal Tokens and obtain the commonly used full-modal basis coordinate system of humans on the basis of the original spatio-temporal organizational form. How to achieve this is the core contribution of our scheme.

Therefore, we believe that artificial intelligence needs to develop to the next stage: the "autonomous interaction" stage. "Autonomous" means that machines are no longer silent "machines". They can spontaneously generate behaviors (which is equivalent to programming themselves). Machines will explore knowledge by themselves (such as actively interacting with the environment to acquire knowledge). "Interaction" means that machines can interact with the environment in real time, update their knowledge in real time, and make continuous decisions to complete complex tasks in unfamiliar environments.

Many renowned scholars have put forward their own views on how to move towards true artificial general intelligence. For example, Professor LeCun proposed the "world model", and Professor Zhu Songchun also proposed four characteristics for achieving artificial general intelligence: (1) being able to execute infinite tasks; (2) being able to autonomously generate new tasks; (3) being driven by a value system; (4) possessing a world model that reflects the real world. Obviously, our scheme is a response to the ideas of Professor LeCun and Professor Zhu Songchun.

“Build a baby machine, then learn lifelong, and grow up”. The idea has been around for many years, now we propose a solution steps with details. We believe that AI technology needs to move on to the next stage: the “autonomous interaction” stage. “Autonomy” means that the machine is no longer a silent machine, it can spontaneously produce behavior (which is equivalent to programming itself). “Interaction” means that the machine can interact with the environment in real time, update its knowledge in real time, and make continuous decisions to complete complex tasks in an unfamiliar environment\cite{ref29}. 

Artificial general intelligence is the original intention of artificial intelligence and also its crowning glory. We have proposed a set of technical schemes for achieving artificial general intelligence, including step-by-step implementation steps. In References \cite{ref25}\cite{ref26}\cite{ref27}\cite{ref28}\cite{ref29}\cite{ref30}\cite{ref31}, we have revealed in detail the technical steps to realize this path in the form of patents. It may be a correct path that guides humans towards artificial general intelligence. In the references, there are articles in which we elaborate on the proposed technical route in more detail.


\end{document}